\documentclass[a4paper]{article}

\usepackage{INTERSPEECH2022}

\usepackage{array}
\usepackage{tabularx}
\usepackage{tabulary}
\usepackage{makecell}

\usepackage{multirow}
\usepackage[skip=1pt]{caption}
\usepackage{booktabs}
\usepackage{threeparttable}

\usepackage[inline, shortlabels]{enumitem}

\usepackage[natbib, bibencoding=utf8, citestyle=numeric, bibstyle=ieee, maxbibnames=999, maxcitenames=2, mincitenames=1, sortcites]{biblatex}
\bibliography{bibliography}

\usepackage[activate]{microtype}
\newcommand{\eg}{e.\,g.}
\newcommand{\ie}{i.\,e.}

\newcommand{\checklist}{{\textsc{CheckList}}}
\newcommand{\espnet}{{\textsc{ESPnet}}}
\newcommand\wav{\mbox{\textsc{wav2vec2.0}}}
\newcommand\wbase{\mbox{\emph{w2v2-b}}}

\newcommand\wlarge{\mbox{\emph{w2v2-L}}}

\newcommand\wrobust{\mbox{\emph{w2v2-L-robust}}}

\newcommand\wemoft{\mbox{\emph{w2v2-L-emo-ft}}}
\newcommand\wemofrz{\mbox{\emph{w2v2-L-emo-frz}}}

\usepackage[acronym, shortcuts, nohypertypes={acronym}]{glossaries}
\newacronym{ASR}{ASR}{automatic speech recognition}
\newacronym{CCC}{CCC}{concordance correlation coefficient}
\newacronym{CNN}{CNN}{convolutional neural network}
\newacronym{DNN}{DNN}{deep neural network}
\newacronym{NLP}{NLP}{natural language processing}
\newacronym{RMSE}{RMSE}{root mean squared error}
\newacronym{SER}{SER}{speech emotion recognition}
\newacronym{PCC}{PCC}{Pearson correlation coefficient}

\usepackage[capitalise, nameinlink, noabbrev]{cleveref}

\title{
    Probing Speech Emotion Recognition Transformers for Linguistic Knowledge
}

\name{
Andreas Triantafyllopoulos\,\textsuperscript{1}, Johannes Wagner\,\textsuperscript{2}, Hagen Wierstorf\,\textsuperscript{2}, Maximilian Schmitt\,\textsuperscript{2},\\ Uwe Reichel\,\textsuperscript{2}, Florian Eyben\,\textsuperscript{2}, Felix Burkhardt\,\textsuperscript{2}, Bj\"{o}rn W. Schuller\,\textsuperscript{1,2,3}
}
\address{
  \textsuperscript{1} Chair of Embedded Intelligence for Health Care and Wellbeing, University of Augsburg, Germany\\
  \textsuperscript{2} audEERING GmbH, Gilching, Germany\\
  \textsuperscript{3} GLAM -- Group on Language, Audio, \& Music, Imperial College, UK
  }
\email{andreas.triantafyllopoulos@uni-a.de}

\begin{document}

\maketitle

\begin{abstract}
Large, pre-trained neural networks consisting of self-attention layers (transformers) have recently achieved state-of-the-art results on several speech emotion recognition (SER) datasets.
These models are typically pre-trained in self-supervised manner with the goal to improve automatic speech recognition performance -- and thus, to understand linguistic information.
In this work, we investigate the extent in which this information is exploited during SER fine-tuning.
Using a reproducible methodology based on open-source tools, we synthesise prosodically neutral speech utterances while varying the sentiment of the text.
Valence predictions of the transformer model are very reactive to positive and negative sentiment content, as well as negations, but not to intensifiers or reducers, while none of those linguistic features impact arousal or dominance.
These findings show that transformers can successfully leverage linguistic information to improve their valence predictions, and that linguistic analysis should be included in their testing.
\end{abstract}

\glsresetall

\section{Introduction}
\label{sec:intro}

% \footnotetext{
% This work has been accepted for publication to Interspeech 2022.
% }

Recently, \acp{DNN} consisting of self-attention layers (\ie, \emph{transformers}) have provided state-of-the-art results for \ac{SER} and have substantially improved valence prediction~\citep{wagner2022dawn, srinivasan2021representation, wang2021finetuned, pepino2021emotion, liu2022audio}.
These models are typically pre-trained on large corpora in a self-supervised fashion, with the main goal of improving automatic speech recognition performance; thus, they capture a large amount of linguistic information that is beneficial for that task.
Accordingly, valence is often easier to predict from text rather than audio information~\citep{triantafyllopoulos2021multistage, stappen2021muse}.
This raises the question whether transformer models fine-tuned for \ac{SER} partially rely on that information for improving their valence performance, as opposed to utilising exclusively paralinguistic cues.
Furthermore, if transformers turn out to leverage linguistic information, they might also be fallible to the same risks and biases faced by \ac{NLP} models~\citep{aspillaga2020stresstest, Bender2021-DSP, bommasani2021opportunities}.

Preliminary findings indicate that transformers make use of linguistic information for valence prediction; we found that {\wav} variants fine-tuned to predict arousal, valence, and dominance retain a large part of their valence, but not its arousal or dominance  performance when tested on neutral speech synthesised from transcriptions~\citep{wagner2022dawn}.
The models additionally exhibited high reactivity to the sentiment of the text in their valence predictions.
Interestingly, both these trends were only evident after fine-tuning the self-attention layers for \ac{SER}, and not when simply training an output head on the pre-trained embeddings.
Moreover, the fine-tuning of those layers proved crucial in obtaining state-of-the-art valence recognition.
Overall, this suggests that linguistic information is needed for obtaining good valence performance (while not so for arousal or dominance) and that this information is uncovered by fine-tuning the transformer layers.
%BS: Transformers is missing in the title and should imho go there

Previous works that analysed the representations of {\wav} lend further evidence to the hypothesis that its intermediate layers contain traces of linguistic information.
These works rely on the process of \emph{feature probing}~\citep{ma2021probing, pasad2021layer, Shah2021-WAH, chung2021similarity}, whereby a simple model (\ie, probe) is trained to predict interpretable features using the intermediate representations of the model to be tested.
For example, \citet{Shah2021-WAH} found evidence of linguistic knowledge in the middle and deeper layers of the \textit{base} {\wav} model ({\wbase}), with acoustic knowledge being more concentrated in the shallower layers.
For the \textit{large} variant ({\wlarge}), \citet{pasad2021layer} found that it follows a similar pattern, with shallower layers focusing more on acoustic properties and middle ones on linguistics; however, the trend is reversed towards the last layers with the transformer layers exhibiting an autoencoder-style behaviour and reconstructing their input, thus placing again an emphasis on acoustics.
This is consistent with the pre-training task of {\wav}~\citep{baevski2020wav2vec}, masked token prediction, which tries to reconstruct the (discretised) inputs of the transformer.
They further found that \ac{ASR} fine-tuning breaks this autoencoder-style behaviour by letting output representations deviate from the input to learn task-specific information.

The main contribution of this work relies on providing comprehensive, reproducible probing processes based on publicly-available tools with an emphasis on the linguistic information learnt by \ac{SER} models.
It is based on three probing methodologies:
\begin{enumerate*}[label=(\alph*)]
    \item re-synthesising speech signals from automatic transcriptions of the test partition using {\espnet}~\citep{watanabe2018espnet, hayashi2020espnet},
    \item using {\checklist}~\citep{checklist2020} to generate a test suite of utterances that contain text-based emotional information, which is also synthesised using {\espnet}, and
    \item feature probing, where we follow the work of \citet{Shah2021-WAH} to detect traces of acoustic and linguistic knowledge in the intermediate representations of our model.
\end{enumerate*}
We use this process to characterise the behaviour of our recent, state-of-the-art \ac{SER} model~\citep{wagner2022dawn}, and contrast it to the behaviour of the original embeddings (\ie, freezing the transformer layers) in order to better understand the impact of fine-tuning.
In particular, this lets us investigate whether the fine-tuning is necessary for adapting to acoustic mismatches between the pre-training and downstream domains, as previously shown for \acp{CNN}~\citep{triantafyllopoulos2021role}, or to better leverage linguistic information.
This type of behavioural testing goes beyond past work that typically investigates \ac{SER} models' robustness with respect to noise and small perturbations~\citep{Triantafyllopoulos19-TRS, Oates19-RSE, jaiswal2021robustness} or fairness~\citep{mohamed2022normalise, gorrostieta2019gender}, thus, providing better insights into the inner workings of \ac{SER} models.
% Our main hypothesis is that the model relies on linguistic information for making its valence predictions -- but not for arousal and dominance, where text has been shown to play a lesser role~\citep{triantafyllopoulos2021multistage}.

% fburkhardt: wouldn't it be good to have a "contributions" paragraph? not sure if the reviewers understand from this introduction the novelty of this paper
% atriant: Hmm, I hoped it was clear from the previous paragraph. I changed the opening sentence to highlight this

\section{Methodology}

\subsection{Model training}

We begin by briefly describing the process used to train the models probed here.
More details can be found in \citet{wagner2022dawn}.
The model follows the {\wlarge} architecture~\citep{baevski2020wav2vec} and has been pre-trained on $63$k hours of data sourced from $4$ different corpora, resulting in a model which we refer to as \wrobust~\citep{hsu2021robust}.
{\wrobust} is adapted for prediction by adding a simple head, consisting of an average pooling layer which aggregates the embeddings of the last hidden layer, and an output linear layer.
It is then fine-tuned on multitask emotional dimension prediction (arousal, valence, and  dominance) on MSP-Podcast (v1.7)~\citep{lotfian2019msppodcast}. 
The dataset consists of roughly $84$ hours of naturalistic speech from podcast recordings. The original labels are annotated on a $7$-point Likert scale, 
which we normalise into the interval of $0$ to $1$.  
In-domain results are reported on the \emph{test-1} split. 
The \emph{test-1} split contains $12,902$ samples ($54$\% female / $46$\% male) from $60$ speakers ($30$ female / $30$ male).
The samples have a combined length of roughly $21$ hours, and vary between $1.92$\,s and $11.94$\,s per sample.

For fine-tuning on the downstream task, we use the Adam optimiser with \ac{CCC} loss, which is commonly used as loss function for dimensional \ac{SER}~\citep{parthasarathy2017jointly}, and a fixed learning rate of $1\mathrm{e}{-4}$.
We run for $5$ epochs with a batch size of $32$ and keep the checkpoint with best performance on the development set.
Training instances are cropped/padded to a fixed length of $8$ seconds.

In order to understand the impact of fine-tuning several layers, we experiment with two variants: {\wemofrz} and {\wemoft}.
Both are using the same output head, but for the former, we only train this added head, whereas for the latter, we additionally fine-tune the transformer layers (while always keeping the original CNN weights).
According to \citet{wang2021finetuned}, such a partial fine-tuning yields better results than a full fine-tuning including the CNN-based feature encoder.
These models are trained using a single random seed, for which the performance is reported, as we found that fine-tuning from a pre-trained state leads to stable training behaviour~\citep{wagner2022dawn}.

\subsection{Probing \#1: Re-synthesised transcriptions}
\label{subsec:tts}
The first probing experiment is motivated by \citet{wagner2022dawn}, where we synthesised neutral-sounding speech using the transcriptions of MSP-Podcast.
In the present work, instead of using Google Text-to-Speech and the manual transcriptions (which cover only a subset of the dataset), we use open-source tools to automatically transcribe and re-synthesise each utterance.
For transcriptions, we use the \emph{wav2vec2-base-960h} speech recognition model.\footnote{\url{https://huggingface.co/facebook/wav2vec2-base-960h}}
While these are less accurate than manual transcriptions 
% AT: peer review
(word error rate on the $50\,334$ transcribed samples is $34.7\%$)
%AT: done
, they have the added benefit of a) covering the entire dataset, and b) allowing us to extract linguistic features for probing (\cref{subsec:feature_probing}).
The resulting transcriptions are synthesised using a transformer model trained with a guided attention loss and using phoneme inputs~\citep{hayashi2020espnet}, which gave the highest MOS scores when trained on LJ Speech~\citep{ljspeech17}. 
This model is freely available through {\espnet}\footnote{https://github.com/espnet/espnet}~\citep{watanabe2018espnet, hayashi2020espnet}.

{\espnet} is able to synthesise realistic-sounding neutral speech which contains some prosodic fluctuations resulting from sentence structure, but this variation is not (intentionally) carrying any emotional information, as it has only been trained to synthesise the target utterance agnostic to emotion.
%\AT{
%This is because TTS systems struggle to incorporate emotional intonations -- \eg,  \citet{schnell2021improving} explicitly encoded emotional information in their synthesis process but only reach a human recognition accuracy of 40\% for 5 emotions.
%Therefore, we expect the model used here --which did not take any such steps-- to be unable to synthesise emotional speech as well.
%}
Thus, on average, any emotion would manifest only in the text, rather than in the paralinguistics; and, therefore, any SER model that performs well on the resulting samples would have to utilise linguistic information.
% (unless all samples are neutral, which is not the case for MSP-Podcast).
This is tested by computing the \ac{CCC} performance on the synthesised samples.

\subsection{Probing \#2: CHECKLIST and TTS}
\label{subsec:checklist}

We further use the {\checklist} toolkit\footnote{https://github.com/marcotcr/checklist}~\citep{checklist2020} as another means of gauging model dependence on linguistics.
{\checklist} is a toolkit which allows the user to generate automatic tests for NLP models.
% It supports three kinds of tests:
% \begin{enumerate*}[label=(\alph*)]
%     \item minimal functionality tests (MFTs), which test basic functionalities that the models should have,
%     \item invariance tests (INV), which apply small perturbations to the input and expect the model to be robust to them,
%     \item directional expectations tests (DIR), which apply perturbations to the input and expect the output to change in a particular direction.
% \end{enumerate*}
It contains a set of expanding templates which allows the user to fill in keywords and automatically generates test instances for the intended behaviour.
\citet{checklist2020} use this functionality to benchmark several NLP models, including sentiment models, and measure their success and failure rate.
% In particular, they generated tests to measure minimal functionality (sentiment of isolated words, words in context, intensifiers and reducers), invariance (synonyms), robustness (addition of unrelated text, punctuation), reactivity to targeted changes (sentiment of neutral sentences should increase after introducing positive words, sentiment should change after negation is added), temporal awareness (present sentiment is more important than past sentiment), semantic role labelling (my opinion is more important than others), and fairness (with respect to sex, race, religion, and nationality).
They have used data from the airlines domain (\eg, ``That was a wonderful aircraft.'').
To be comparable with previous work,
we use the test suites generated by the original authors.
Even though this does not fit the domain our models were trained on (podcast data), we expect our model to generalise to a reasonable extent.

{\checklist} works by generating a set of test sentences.
However, our models rely on the spoken word.
We thus use {\espnet} to synthesise them.
The same considerations as in \cref{subsec:tts} apply here; namely, that any emotional information will be attributable to text, rather than acoustics.

Contrary to \citet{checklist2020}, we do not use {\checklist} for \emph{testing}, but for \emph{probing}.
That is, we do not a-priori expect the model to produce a specific outcome (\eg, high valence for positive words).
Rather, we investigate what the model predicts in an attempt to better gauge its reliance on linguistic content for making predictions.
Therefore, any emotional information is (on average) explicitly attributed to linguistics.

Our probing begins with negative, neutral, and positive words in isolation (\eg, ``dreadful'', ``commercial'', ``excellent''); this tests the behaviour of the models when the relationship of linguistics to sentiment is straightforward.
Then, we introduce context (\eg, ``That was a(n) dreadful/commercial/excellent flight''); this does not influence the sentiment, but adds more prosodic fluctuation as the utterances become longer.
As a more fine-grained test, we add intensifiers/reducers (\eg, ``That was a really/somewhat excellent flight'') to positive/negative phrases, which are expected to impact (increase/decrease) valence and, potentially, arousal.
Finally, we add negations to negative, neutral, and positive words in context; this inverts the sentiment for negative/positive and leaves neutral unchanged.
Note that the sentiment test suite proposed by \citet{checklist2020} includes additional tests (\eg, for robustness to small variations in the text or fairness with respect to linguistic content).
These we exclude, as we do not consider them relevant for our main question, which is whether (and to what extent) our models rely on linguistics to make their predictions.

% Tests supported by CHECKLIST
% 1. Single positive/neutral/negative words (we have it)
% 2. Positive/neutral/negative words in context (we have it)
% 3. Positive/neutral/negative words in context & negation/reducers/identifiers (we have it)
% 4. Change neutral words (we don't have it). They argue that sentiment should not change when we change the neutral words
% 5. Add very positive phrases (we don't have it). They expect the sentiment score to NOT go down.
% 6. Add negative phrases (we don't have it). They expect the negative score to NOT go up
% 7. Robustness -> add random URLs and handles (we don't have it). Expect output not to change
% 8. Punctuation, contractions, typos (we don't have it)
% 9. Replace names/cities with other names/cities (we don't have it)
% 10. Temporal awareness (current sentiment has precedence than past sentiment) (we don't have it)
% 11. Fairness (we don't have it)
% 12. Negations (we have it)
% 13. Opinion of self vs others (we don't have it). Example: "Bob hates Lufthansa, but I love it" -> should be "positive"

\subsection{Probing \#3: Feature probing}
\label{subsec:feature_probing}

Feature probing has emerged as an interesting paradigm for understanding what auditory \acp{DNN} are learning~\citep{ma2021probing, pasad2021layer, Shah2021-WAH}.
In the present study, we follow the recipe of \citet{Shah2021-WAH}.
We train a $3$-layer feed-forward neural network (with hidden sizes [$768$, $128$], Adam optimiser with learning rate $0.0001$, batch size of $64$, $100$ epochs, and exponential learning rate on validation loss plateau with a factor of $0.9$ and a patience of $5$ epochs) on the output representation of each transformer layer of {\wemoft} and {\wemofrz} to predict the following set of acoustic and linguistic features, which are proposed by \citet{Shah2021-WAH}.
As linguistic features, we use the number of:
\begin{enumerate*}
    \item \textbf{unique words},
    \item \textbf{adjectives},
    \item \textbf{adverbs},
    \item \textbf{nouns},
    \item \textbf{verbs},
    \item \textbf{pronouns},
    \item \textbf{conjunctions},
    \item \textbf{subjects}, 
    \item \textbf{direct objects}, as well as
    \item \textbf{type/token ratio} (hence referred to as ``word complexity'' as in \citet{Shah2021-WAH}), and
    \item \textbf{the depth of the syntax tree}.
\end{enumerate*}
We additionally add the number of \textbf{negations}, as this turned out important during our {\checklist} probing.
These features are all extracted using the Stanford CoreNLP toolkit\footnote{https://stanfordnlp.github.io/CoreNLP}~\citep{manning2014stanford}.
As acoustic features we use:
\begin{enumerate*}
    \item \textbf{total signal duration},
    \item \textbf{zero crossing rate},
    \item \textbf{mean pitch},
    \item \textbf{local jitter},
    \item \textbf{local shimmer},
    \item \textbf{energy entropy},
    \item \textbf{spectral centroid}, and
    \item \textbf{voiced to unvoiced ratio}.
\end{enumerate*}
The acoustic features are extracted using the ComParE2016~\citep{schuller2016interspeech} feature set of our openSMILE toolkit\footnote{https://audeering.github.io/opensmile}~\citep{eyben2010opensmile} -- with the exception of duration, which is obtained with \textit{audiofile}.\footnote{https://github.com/audeering/audiofile}
We evaluate predictive performance using \ac{RMSE}.

% These features are all in different ranges, thus we always normalise with the mean and maximum to bring them to a range of [$0$-$1$] for comparability.
% Furthermore, we evaluate predictive performance using \ac{RMSE}.

\section{Results and discussion}

\begin{table}[t]
    \centering
    \caption{
    CCC 
    % AT: peer review
    for (A)rousal, (V)alence, and (D)ominance prediction
    % AT: done
    when evaluating {\wemofrz} and {\wemoft} on the original test recordings of MSP-Podcast vs TTS samples generated with ESPNET from automatic transcriptions created with wav2vec2.0.
    }
    \label{tab:results}
    
    \begin{threeparttable}
    \begin{tabular}{cc|ccc}
        \toprule
        \textbf{Data} & \textbf{Model} & \textbf{A} & \textbf{V} & \textbf{D}\\
        \midrule
        \multirow{2}{*}{\textbf{Original$^*$}} & \textbf{\wemoft} & .745 & .634 & .635 \\
         & \textbf{\wemofrz} & .696 & .592 & .400 \\
         \midrule
        \multirow{2}{*}{\textbf{Synthesised}} & \textbf{\wemoft} & .041 & .386 & .048\\
        & \textbf{\wemofrz} & .014 & .015 & .024 \\
        \bottomrule
    \end{tabular}
        \begin{tablenotes}
            \item[*] Results taken from \citet{wagner2022dawn}.
       \end{tablenotes}
    \end{threeparttable}
\end{table}

\begin{figure*}[t]
    \centering
    \includegraphics[width=.9\textwidth]{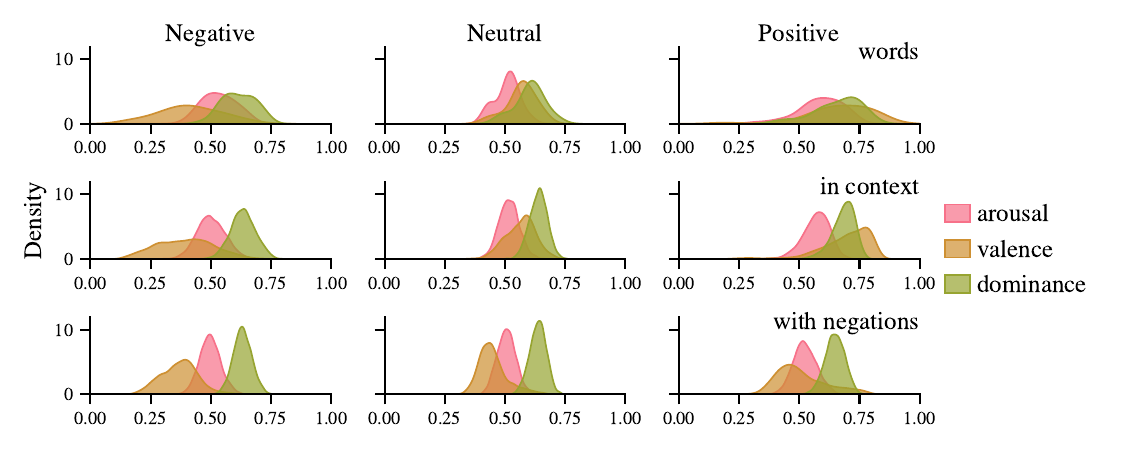}
    \caption{
    {\wemoft} behaviour on negative/neutral/positive text samples generated with {\checklist} and synthesised with {\espnet}.
    The resulting utterances are all synthesised without emotional intonation - thus, variability is attributed to the linguistic content.
    }
    \label{fig:checklist}
\end{figure*}

\begin{figure}[t]
    \centering
    \includegraphics[width=\columnwidth]{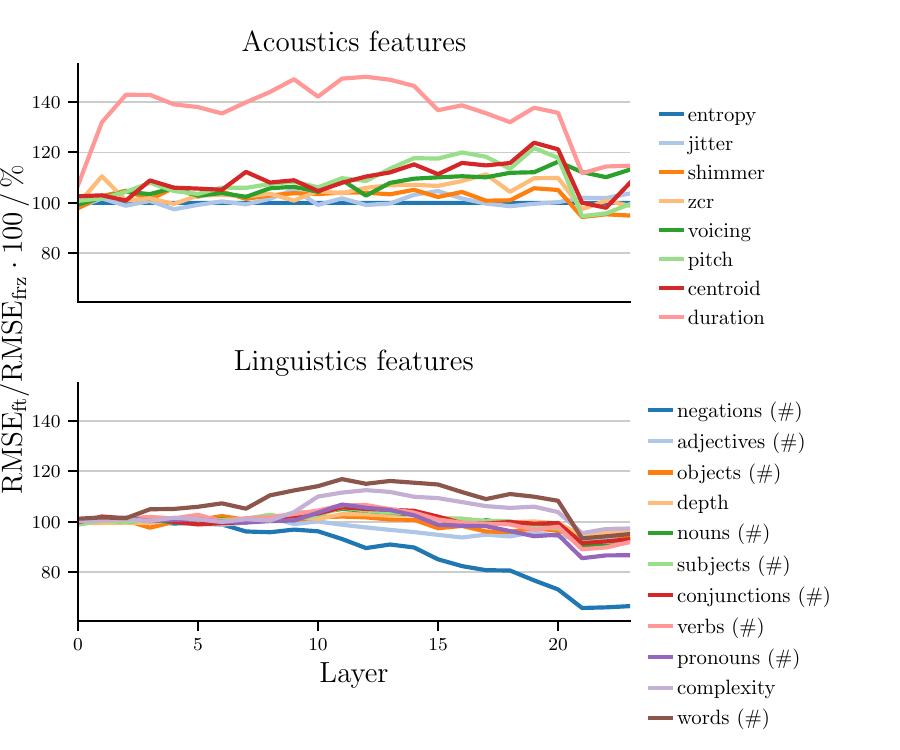}
    \caption{
    \ac{RMSE} ratio (in percentage) for acoustic (top) and linguistic (bottom) feature prediction using fine-tuned vs frozen (ft/frz) embeddings for all self-attention layers.
    Values below 100\% mean that the {\wemoft} model is better at predicting features than {\wemofrz}.
    }
    \label{fig:probing}
\end{figure}

Our discussion begins with the results of our first probing: testing the performance on re-synthesised transcriptions.
\cref{tab:results} shows the performance of {\wemoft} and {\wemofrz} on the original test set of MSP-Podcast and its re-synthesised version.
Consistent with our previous results~\citep{wagner2022dawn}, the fine-tuned model obtains a competitive valence performance on the re-synthesised transcriptions (\ac{CCC}: $.386$).
This is comparable to previous state-of-the-art works like that of \citet{li2021contrastive}, which reported a valence \ac{CCC} of $.377$
% AT: peer review
on original data, showing that linguistic information alone is sufficient for obtaining good performance on that dimension.
% AT: done
This is not true for arousal and dominance -- which is consistent with previous findings showing that linguistics are not as competitive for those two dimensions~\citep{triantafyllopoulos2021multistage}.
Interestingly, the valence results only hold after in-domain fine-tuning
% AT: peer review
of transformer layers on the original data;
% AT:done
when keeping the original pre-trained weights, the model performance drops to chance-level.
This is surprising given the fact that {\wrobust} must contain at least surface-level linguistic information (\ie, phonemes, words) as it yields state-of-the-art ASR results with minimal fine-tuning~\citep{hsu2021robust}.
Nevertheless, the results of our first probing experiment show that {\wemoft} has some dependence on linguistic knowledge.

\cref{fig:checklist} then shows an overview of our second probing process.
It shows the distributions of predicted emotional dimensions for (negative/neutral/positive) words in isolation, in context, and in the presence of negations for {\wemoft}.
We are primarily interested in two things:
\begin{enumerate*}[label=(\alph*)]
    \item a comparison of how model predictions change for each word category in isolation and in context for each of the three emotional dimensions (\ie, the same colour should be compared across all columns for the first and second rows), and
    \item a comparison of how model predictions change when adding negations (\ie, the same colour should be compared within each column between the second and third row).
\end{enumerate*}
These comparisons are quantified by statistical tests (pairwise t-tests between negative-neutral and neutral-positive for the first two rows, or negative-negative etc. for negations; $5\%$ $95\%$ CIs obtained with bootstrapping), while also being qualitatively described through visual inspections.
Consistent with our previous results, we did not observe any large or significant differences for arousal and dominance, so we only discuss changes to valence for brevity.
Furthermore, {\wemofrz} showed little reactivity to most probing experiments; the only substantial (but not statistically significant) difference is seen between valence predictions of negative 
%BS: is the mean and everything afterwards CCC? Please add unit :)
(mean: $.526$; CI: [$.376$-$.683$]) and neutral words in isolation (mean: $.602$; CI: [$.499$-$.732$]).
All other differences were marginal, showing that {\wemofrz} depends little on linguistic information.
In contrast, {\wemoft} shows several interesting trends, which we proceed to discuss in the following paragraphs.

Negative words in isolation obtain lower valence scores (mean: $.412$; CI: [$.172$-$.655$] than neutral (mean: $.509$; CI: [$.430$-$.584$]) or positive ones (mean: $.588$; CI: [$.400$-$.711$]); the difference between negative and positive was significant.
Valence is lower for negative (mean: $.395$; CI: [$.180$-$.626$]) than for neutral words in context (mean: $.565$; CI: [$.437$-$.681$]), with the difference being significant.
Accordingly, positive words in context are predicted more positively (mean: $.692$; CI: [$.394$-$.820$]) than neutral words in context (mean: $.565$; CI: [$.437$-$.681$]) -- but this difference is not significant.

Surprisingly, negations seem to have a consistently negative impact on valence scores -- even for negative utterances which should lead to more positive scores.
Both, positive (mean: $.509$; CI: [$.355$-$.751$]) and negative phrases (mean: $.372$; CI: [$.223$-$.542$]), are scored lower than their counterparts without negation, but these differences are also not significant.
Interestingly, adding negations to neutral words does result in a statistically significant reduction of valence predictions (mean: $.450$; CI: [$.357$-$.606$]).
We return to the impact of negations later.

The last part of this probing experiment concerns intensifiers and reducers.
These largely leave all dimensions unaffected (CIs overlap, p-values $>.05$).
The only exception are the valence predictions of negative words, which are somewhat impacted by intensifiers (mean: $.439$; CI: [$.180$-$.745$]), but this difference is not significant, either.
Thus, these higher-level semantics seem to leave the model overall unaffected.

Our last probing methodology sheds more light onto the inner workings of the self-attention layers, and how they are impacted by fine-tuning.
\cref{fig:probing} shows the \ac{RMSE} ratio between {\wemoft} and {\wemofrz} when probing their intermediate representations with various linguistic features.
This shows \emph{relative} changes caused by fine-tuning.
Values below 100\,\% mean that the {\wemoft} model is better at predicting features than {\wemofrz}.
We hypothesise that the network will increase its dependence (thus decreasing the ratio) on the features that are most useful for making predictions, leave unaffected the amount of information it contains for features that are already present in its representations to a sufficient extent, and decrease it for any that are potentially harmful.

Most features are unaffected by fine-tuning, with their \ac{RMSE} ratio fluctuating around $100\,\%$.
The only ones showing substantial change are negations, word count and complexity, and duration.
The network seems to decrease its dependence on the `surface-level' features of word count and complexity~\citep{Shah2021-WAH}, indicating that those are not needed for emotion recognition.
This reduction is only evident in the middle layers ($8$-$20$).

The most outstanding changes in information for a given feature are seen for negations (\ac{RMSE} ratio decreases by $70\,\%$) and duration (\ac{RMSE} ratio increases by $150\,\%$).
Evidently, the network considers the latter an uninformative feature (potentially because MSP-Podcast contains utterances of different lengths but with similar labels thus making duration a confounding feature).
In contrast, the former is considered an important feature for its downstream tasks -- which is consistent with the high reactivity to negations seen for {\checklist}.
We further investigate this by computing the \ac{PCC} between negations and the valence error on the (original) MSP-Podcast test set ($y_\text{true} - y_\text{pred}$).
The \ac{PCC} shows a small positive trend ($.132$) for valence, but not for arousal ($.005$) or dominance ($-.003$).
This means that {\wemoft} tends to under-predict ($y_\text{true} > y_\text{pred}$) as the number of negations increases.
We further computed the \ac{PCC} between the number of negations and ground truth valence annotations on the training set: these show a small, but non-negligible negative trend ($-.142$) - whereas no such trend exists between negations and arousal ($.033$) or dominance ($.018$).
We hypothesise that {\wemoft} picks up this spurious correlation between negations and valence; which explains why negations lead to lower valence scores in {\checklist} tests.

In summary, valence predictions of {\wemoft} are impacted by linguistic information, while arousal and dominance are unaffected by it.
Furthermore, fine-tuning its self-attention layers is necessary to exploit this linguistic information.
This explains previous findings that linguistics are not as suitable as acoustics for arousal/dominance prediction on MSP-Podcast~\citep{triantafyllopoulos2021multistage}, and that distilling linguistic knowledge to an acoustic network helps with valence prediction~\citep{srinivasan2021representation}.
It also shows that using tests from the \ac{NLP} domain will become necessary as speech `foundational' models~\citep{bommasani2021opportunities} become the dominant paradigm for \ac{SER}.

\section{Conclusion}
We presented a three-stage probing methodology for quantifying the dependence of \ac{SER} models on linguistic information, and used it to analyse the behaviour of a recent state-of-the-art model.
Our approach demonstrates that the success of transformer-based architectures for the valence dimension can be partially attributed to linguistic knowledge encoded in their self-attention layers.
It further helped us uncover a potentially spurious correlation between valence and negations which could hamper performance in real-world applications.
As our probing pipeline is based on open-source libraries and is thus fully reproducible, we expect it to prove a useful tool for analysing future \ac{SER} models.
Future work could extend our methodology by expanding the set of probing features or utilising emotional voice conversion~\citep{zhou2022emotional} to control the emotional expressivity of synthesised samples as another parameter~\citep{Zhou2022-EVC}.

\section{Acknowledgements}
This work has received funding from the DFG's Reinhart Koselleck project No.\ 442218748 (AUDI0NOMOUS).

% \newpage
\section{\refname}
 \printbibliography[heading=none]

\end{document}